\documentclass[journal,transmag]{IEEEtran}
\usepackage{cite}
\usepackage{amsmath,amssymb,amsfonts}
\usepackage{algorithmic}
\usepackage{graphicx}
\usepackage{textcomp}
\usepackage{xcolor}
\usepackage{multicol}
\usepackage{multirow}
\usepackage{fontawesome}
\usepackage[overload]{empheq}

\ifCLASSINFOpdf
 
\else
 
\fi

\begin{document}
 
\title{A Comprehensive Survey on Transformers for Computer Vision Applications}

\author{\IEEEauthorblockN{Sonain Jamil\IEEEauthorrefmark{1}, 
Md. Jalil Piran\IEEEauthorrefmark{2},~\IEEEmembership{Senior Member,~IEEE} and Oh-Jin Kwon\IEEEauthorrefmark{1}}
\IEEEauthorblockA{\IEEEauthorrefmark{1}Department of Electronics Engineering, Sejong University, Seoul 05006, Korea}
\IEEEauthorblockA{\IEEEauthorrefmark{2}Department of Computer Science Engineering, Sejong University, Seoul 05006, Korea}
\thanks{Manuscript received Month xx, 2022; revised Month xx, 2022. 
Corresponding author: M. J. Piran (email: piran@sejong.ac.kr) and O.-J. Kwon (email: ojkwon@sejong.ac.kr).}}

\markboth{Journal , Vol. xx, No. x, April~2022}%
{S. Jamil \MakeLowercase{\textit{et al.}}: A Comprehensive Survey of Transformers for Computer Vision}

\IEEEtitleabstractindextext{%
\begin{abstract}
As a special type of transformer, Vision Transformers (ViTs) are used to various computer vision applications (CV), such as image recognition. There are several potential problems with convolutional neural networks (CNNs) that can be solved with ViTs. For image coding tasks like compression, super-resolution, segmentation, and denoising, different variants of the ViTs are used. The purpose of this survey is to present the first application of ViTs in CV. The survey is the first of its kind on ViTs for CVs to the best of our knowledge. In the first step, we classify different CV applications where ViTs are applicable. CV applications include image classification, object detection, image segmentation, image compression, image super-resolution, image denoising, and anomaly detection. Our next step is to review the state-of-the-art in each category and list the available models. Following that, we present a detailed analysis and comparison of each model and list its pros and cons. After that, we present our insights and lessons learned for each category. Moreover, we discuss several open research challenges and future research directions. 
\end{abstract}

\begin{IEEEkeywords}
Vision Transformers, Computer Vision, Deep learning, Image coding.
\end{IEEEkeywords}}

\maketitle

\IEEEdisplaynontitleabstractindextext
 
\IEEEpeerreviewmaketitle

\section{Introduction}
\IEEEPARstart{V}{ision} transformers (ViTs) are designed for tasks related to vision, including image recognition~\cite{B.Heo2021}. Originally, transformers were used to process natural language (NLP). Bidirectional encoder representations from transformers (BERT)~\cite{BERT} and generative pre-trained transformer 3 (GPT-3)~\cite{GPT-3} were the pioneers of transformer models for natural language processing. In contrast, classical image processing systems use convolutional neural networks (CNNs) for different computer vision (CV) tasks. The most common CNN models are AlexNet \cite{AlexNet}, ResNet \cite{ResNet}, VGG \cite{VGG}, GoogleNet \cite{GoogleNet}, Xception \cite{Xception}, Inception \cite{Inception}, DenseNet \cite{DenseNet}, and EfficientNet \cite{EfficientNet}.

To track attention links between two input tokens, transformers are used. With an increasing number of tokens, the cost rises inexorably. The pixel is the most basic unit of measurement in photography while calculating every pixel relationship in a normal image would be time-consuming and memory-intensive. ViTs, however, take several steps as described below.
\begin{enumerate}
    \item ViTs divide the full image into a grid of small image patches.
    \item ViTs apply linear projection to embed each patch.
    \item Then, each embedded patch becomes a token, and the resulting sequence of embedded patches is passed to the transformer encoder (TE). 
    \item Then, TE encodes the input patches, and the output is given to the multi-layer perceptron (MLP) head, and the output of the MLP head is the input class.
\end{enumerate}

\begin{figure*}
	\includegraphics[width=\textwidth]{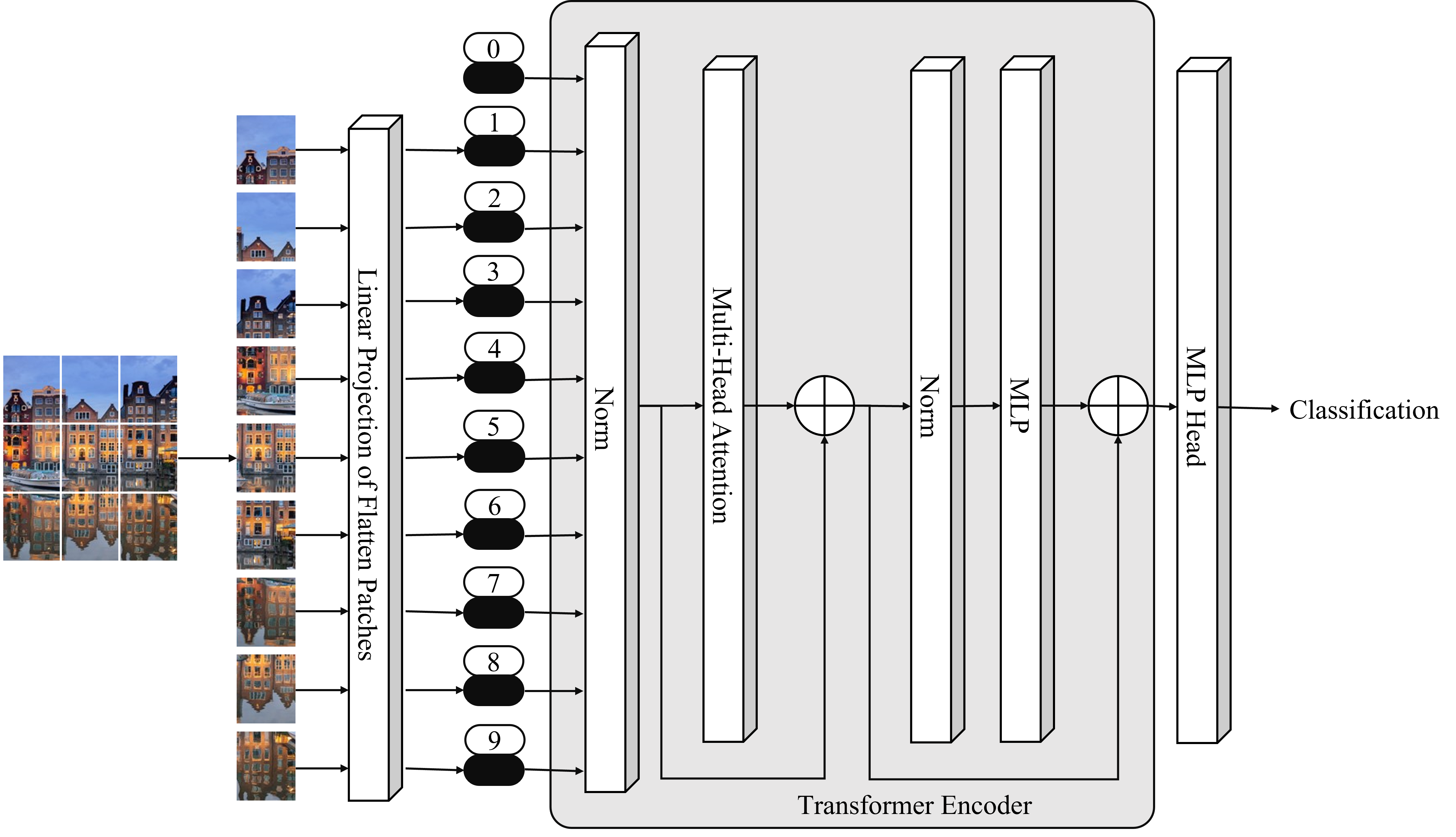}
	\caption{ViT for Image Classification.}
	\label{ViT}
\end{figure*}

Figure \ref{ViT} shows the primary illustration of ViTs. In the beginning, the input image is divided into smaller patches. Each patch is then embedded using linear projection. Tokens are created from embedded patches that are given to the TE as inputs. Multi-head attention and normalization are used by TE to encode the information embedded in patches. TE output is given to the MLP head, and MLP head output is the input image class.

For image classification, the most popular architecture uses the TE to convert multiple input tokens. However, the transformer's decoder can also be used for other purposes.
As described in 2017, transformers have rapidly spread across NLP, becoming one of the most widely used and promising designs~\cite{Vaswani2017}. 

For CV tasks, ViTs were applied in 2021~\cite{Dosovitskiy2020}. The aim was to construct a sequence of patches that, once reconstructed into vectors, are interpreted as words by a standard transformer. Imagine that the attention mechanism of NLP transformers was designed to capture the relationships between different words within the text. In this case, CV takes into account how the different patches of the image relate to one another. 

In 2021, a pure transformer outperformed CNNs in image classification \cite{Dosovitskiy2020}. In June 2021, a transformer backend was added to the conventional ResNet, drastically lowering costs while enhancing accuracy \cite{Wu2020},\cite{Xiao2021}.

In the same year, several key ViTs versions were released. Various variants were more efficient, accurate, or applicable to specific regions. Swin transformers are the most prominent variants \cite{Liu2021}. Using a multi-stage approach and altering the attention mechanism, the Swin transformer achieved cutting-edge performance on object detection datasets. There is also the TimeSformer, which was proposed for video comprehension issues and may capture spatial and temporal information through divided space-time attention~\cite{Bertasius2021}.

ViTs performance is influenced by decisions like optimizers, dataset-specific hyperparameters, and network depth. Optimizing CNN is significantly easier. 

Even when trained on data quantities that are not as large as those required by ViTs, CNNs perform admirably.

Apparently, CNNs exhibit this distinct behavior because of some inductive biases that they can use to comprehend the particularities of images more rapidly, even if they end up restricting them, making it more difficult for them to recognize global connections. ViTs, on the other hand, are devoid of these biases, allowing them to capture a broader and more global set of relationships at the expense of more difficult data training \cite{Raghu2021}.

ViTs are also more resistant to input visual distortions such as hostile patches and permutations \cite{Naseer2021}. Conversely, preferring one architecture over another may not be the best choice. The combination of convolutional layers with ViTs has been shown to yield excellent results in numerous CV tasks \cite{Dai2021}-\cite{Coccomini2021}.

To train these models, alternate approaches were developed due to the massive amount of data required. It is feasible to train a neural network virtually autonomously, allowing it to infer the characteristics of a given issue without requiring a large dataset or precise labeling. It might be the ability to train ViTs without a massive vision dataset that makes this novel architecture so appealing.

ViTs have been employed in numerous CV jobs with outstanding and, in some cases, cutting-edge outcomes.
The following are some of the important application areas:
\begin{itemize}
    \item Image Classification
    \item Anomaly Detection
    \item Object Detection
    \item Image Compression
    \item Image Segmentation
    \item Video Deepfake Detection
    \item Cluster Analysis
\end{itemize}

\begin{figure}
	\includegraphics[width=0.45\textwidth]{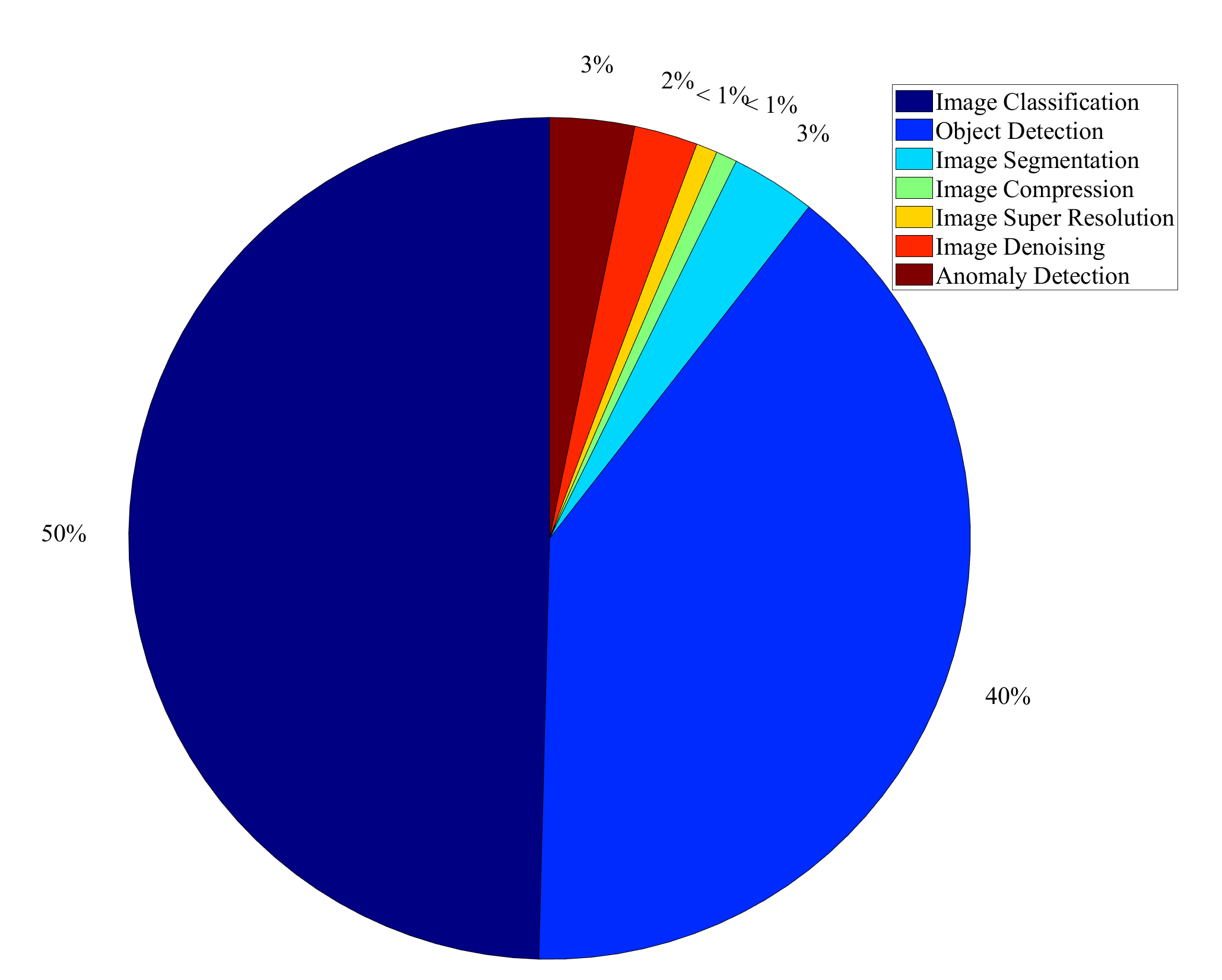}
	\caption{Use of ViTs for CV applications.}
	\label{plot}
\end{figure}

Figure \ref{plot} shows that the percentage of the application of ViTs for image classification, object detection, image segmentation, image compression, image super resolution, image denoising and anomaly detection is 50\%, 40\%, 3\%, less than 1\%, less than 1\%, 2\% and 3\% respectively.
\begin{figure*}[t]
	\includegraphics[width=\textwidth]{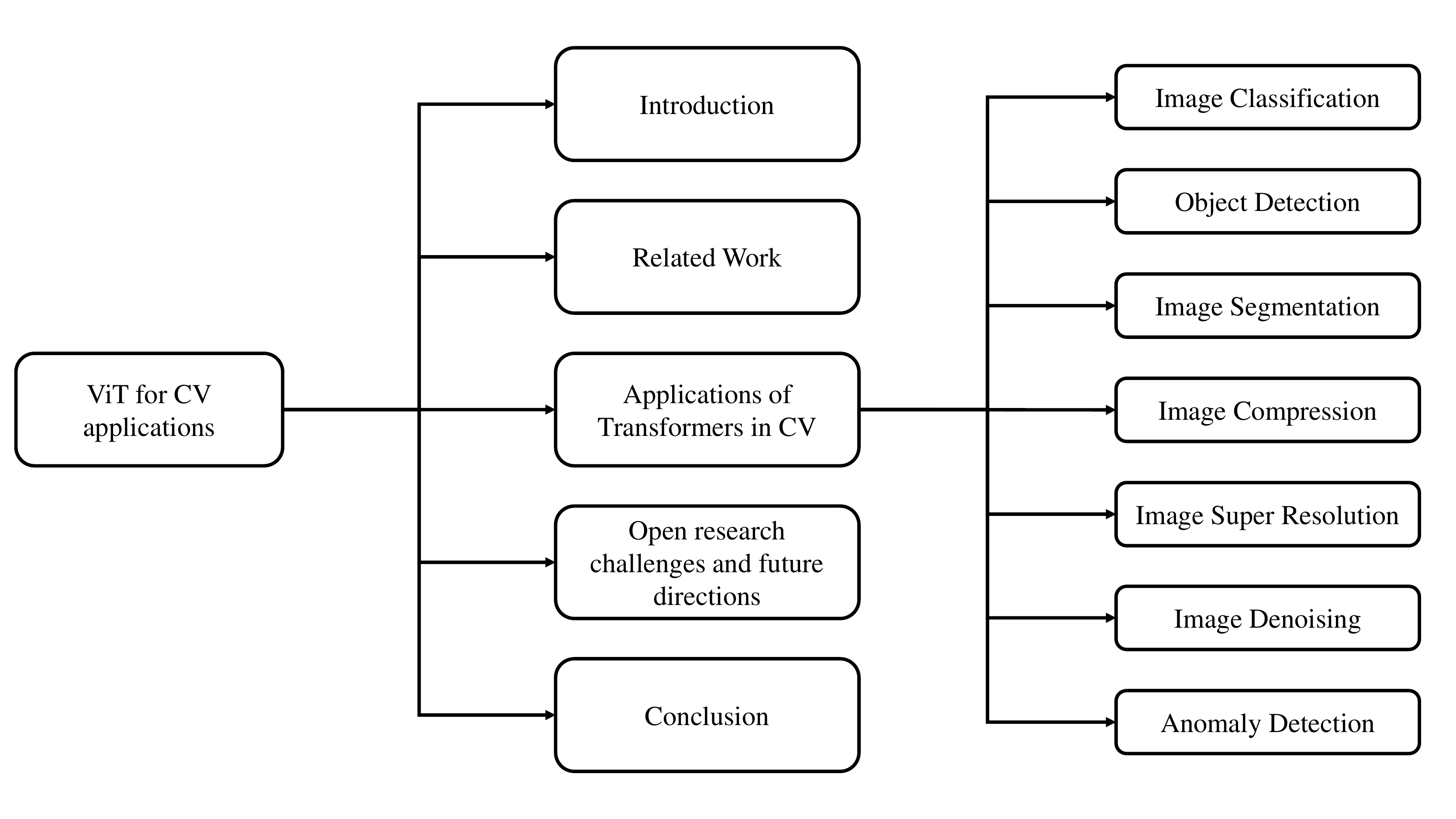}
	\caption{Organization of the survey.}
	\label{Organization}
\end{figure*}

ViTs have been widely utilized in CV tasks. ViTs can solve the problems faced by the CNNs. Different variants of ViTs are used for the image compression, super-resolution, denoising, and segmentation. With the advancement in the ViTs for CV applications, a state-of-the-art survey is required which highlights the performance of ViTs for the CV tasks. In this survey, we first classify different applications of CV such as image classification, object detection, image segmentation, image compression, image super-resolution, image denoising, and anomaly detection where ViTs are used. In the next step we survey the state-of-the-art in each CV application and tabulate the existing ViT-based models. We also discuss the pros and cons of each listed model. We present the lessons learnt for each CV application. In the end, we discuss several open research issues and future directions.

Following is a breakdown of the survey. The upcoming section presents related work, while Section 3 presents applications of ViTs in CV. In section 4, we present open research challenges and future directions. Lastly, Section 5 concludes the survey. Figure \ref{Organization} shows the organization of the survey.

\begin{table}[t]
    \centering
     \caption{List of Acronyms}
    \begin{tabular}{p{50pt} p{170pt}}
    \hline
    \hline
    \textbf{Acronym}& 
    \textbf{Meaning}\\
    \hline
    \hline
    AID& 
    Aerial image dataset\\
    AP&
    Average precision\\
    AQI&
    Air quality index\\
    AUC&
    Area under the curve\\
    AUROC&
    Area under receiver operating characteristic curve\\
    $AP^{box}$&
    Box average precision\\
    BERT&
    Bidirectional encoder representations from transformers\\
    bpp&
    Bits per pixel\\
    BrT&
    Bridged transformer\\
    BTAD&
    BeanTech anomaly detection\\
    CIFAR&
    Canadian institute for advanced research\\
    CV& 
    Computer vision\\
    CNN& 
    Convolutional neural network\\ 
    DHViT&
    Deep hierarchical ViT\\
    DOViT&
    Double output ViT\\
    ES-GSNet&
    Excellent teacher guiding small networks \\
    GAOs-1&
    Get AQI in one shot-1\\
    GAOs-2&
    Get AQI in one shot-2\\
    GPT-3&
    Generative pre-trained transformer 3\\
    HD&
    Hausdorff distance\\
    IoU&
    Intersection over union\\
    JI&
    Jaccord index\\
    LiDAR&
    Light detection and ranging\\
    mAP&
    Mean average precision\\
    MLP&
    Multi layer perceptron\\
    MIL-ViT&
    Multiple instance enhanced ViT\\ 
    MIM&
    Masked image modeling\\
    MITformer&
    Multi-instance ViT\\
    MSE&
    Mean squared error\\
    MS-SSIM&
    Multi-scale structural similarity\\
    NLP&
    Natural language processing\\
    NWPU&
    Northwestern Polytechnical University\\
    PSNR&
    Peak signal to noise ratio\\
    PRO&
    Per region overlap\\
    PUAS&
    Planet understanding the amazon from space\\
    RFMiD2020&
    2020 retinal fundus multi-disease image datase\\
    R-CNN&
    Region-based convolutional neural network\\
    RMSE&
    Root mean square error\\
    SSIM&
    Structural similarity\\
    TE&
    Transformer encoder\\
    UCM&
    UC-Mered land use dataset\\
    ViTs& 
    Vision transformers\\
    VT-ADL&
    ViT network for image anomaly detection and localization\\
    ViT-PP&
    ViT with post processing\\
    YOLOS&
    You only look at one sequence\\
    \hline 
    \hline
    \end{tabular}
    \label{TableAbb}
\end{table}

\section{Related Work}
A number of surveys have been conducted on ViTs in the literature.~\cite{Tay2020} reviews the theoretical concepts, foundation, and applications of the transformer for memory efficiency. They also discussed the applications of efficient transformers in NLP. CV tasks, however, were not included. A similar study,~\cite{SKhan2021}, examined the theoretical aspects of the ViTs, the foundations of transformers, the role of multi-head attention in transformers, and applications of transformers in image classification, segmentation, super-resolution, and object detection. The survey did not include applications of transformers for image denoising and compression. 

In \cite{Liu2021S}, the authors described the architectures of transformers for segmenting, classifying, and detecting objects in images. This survey did not include tasks such as image super-resolution, denoising, and compression associated with CV and image processing. 

Lin {\em et al.} in \cite{Lin2021} summarized different architectures of NLP. The survey, however, did not include any applications of transformers for CV tasks.
In~\cite{Xu2022}, the authors discuss different architectures of transformers for computational visual media. The authors discussed the application of transformers for low-level vision and generation, such as image colorization, image super-resolution, image generation, and text-to-image conversion. Additionally, the survey focused on high-level vision tasks such as segmentation and object detection. However, the survey did not discuss the transformer for image compression and classification.

Table \ref{TableSummary} summarizes all existing surveys on the ViTs. As a result of an analysis of Table \ref{TableSummary}, it is evident that the survey is needed to provide insight into the latest developments in ViTs for several image processing and CV tasks, including classification, detection, segmentation, compression, denoising, and super resolution. 

\begin{table*}
    \centering
     \caption{Summary of the available surveys on ViTs.}
    \begin{tabular}{p{20pt} p{20pt} p{40pt} p{40pt} p{40pt} p{40pt} p{40pt} p{40pt} p{40pt} p{80pt}}
    \hline
    \hline
         \multirow{2}{20pt}{\textbf{Survey}}& 
         \multirow{2}{20pt}{\textbf{Year}}& 
         \multicolumn{7}{p{240pt}} {\centering \textbf{Scope}}&
         \multirow{2}{80pt}{\textbf{Contributions and limitations}}\\\cline{3-9}
         & 
         & 
         \textbf{Image Classification}& 
         \textbf{Object Detection}& 
         \textbf{Image Segmentation}& 
         \textbf{Image Compression}& 
         \textbf{Image Super Resolution}& 
         \textbf{Image Denoising}&
         \textbf{Anomaly Detection}&
         \\ 
         \hline
         \vspace{0.01cm} \cite{Tay2020} \vspace{0.1cm}& 
         \vspace{0.01cm} 2020 \vspace{0.1cm}& 
         \vspace{0.01cm} \textcolor{red}{\faTimesCircle} \vspace{0.1cm}& 
         \vspace{0.01cm} \textcolor{red}{\faTimesCircle} \vspace{0.1cm}& 
         \vspace{0.01cm} \textcolor{red}{\faTimesCircle} \vspace{0.1cm}& 
         \vspace{0.01cm} \textcolor{red}{\faTimesCircle} \vspace{0.1cm}& 
         \vspace{0.01cm} \textcolor{red}{\faTimesCircle} \vspace{0.1cm}&
         \vspace{0.01cm} \textcolor{red}{\faTimesCircle} \vspace{0.1cm}&
         \vspace{0.01cm} \textcolor{red}{\faTimesCircle} \vspace{0.1cm}&
         \vspace{0.01cm} Survey of foundation and applications of efficient transformers \vspace{0.1cm}\\
         \hline
         \vspace{0.01cm} \cite{SKhan2021} \vspace{0.1cm}& 
         \vspace{0.01cm} 2021 \vspace{0.1cm}& 
         \vspace{0.01cm} \textcolor{green}{\faCheckCircle} \vspace{0.1cm}& 
         \vspace{0.01cm} \textcolor{green}{\faCheckCircle} \vspace{0.1cm}& 
         \vspace{0.01cm} \textcolor{green}{\faCheckCircle} \vspace{0.1cm}& 
         \vspace{0.01cm} \textcolor{red}{\faTimesCircle} \vspace{0.1cm}& 
         \vspace{0.01cm} \textcolor{green}{\faCheckCircle} \vspace{0.1cm}& 
         \vspace{0.01cm} \textcolor{red}{\faTimesCircle} \vspace{0.1cm}&
         \vspace{0.01cm} \textcolor{red}{\faTimesCircle} \vspace{0.1cm}&
         \vspace{0.01cm} Survey of basic concepts and applications of transformers in CV \vspace{0.1cm}\\
         \hline
         \vspace{0.01cm} \cite{Lin2021} \vspace{0.1cm}& 
         \vspace{0.01cm} 2021 \vspace{0.1cm}& 
         \vspace{0.01cm} \textcolor{red}{\faTimesCircle} \vspace{0.1cm}& 
         \vspace{0.01cm} \textcolor{red}{\faTimesCircle} \vspace{0.1cm}& 
         \vspace{0.01cm} \textcolor{red}{\faTimesCircle} \vspace{0.1cm}& 
         \vspace{0.01cm} \textcolor{red}{\faTimesCircle} \vspace{0.1cm}& 
         \vspace{0.01cm} \textcolor{red}{\faTimesCircle} \vspace{0.1cm}& 
         \vspace{0.01cm} \textcolor{red}{\faTimesCircle} \vspace{0.1cm}&
         \vspace{0.01cm} \textcolor{red}{\faTimesCircle} \vspace{0.1cm}&
         \vspace{0.01cm} Survey of different architectures of transformers \vspace{0.1cm}\\
         \hline
         \vspace{0.01cm} \cite{Liu2021S} \vspace{0.1cm}& 
         \vspace{0.01cm} 2021 \vspace{0.1cm}& 
         \vspace{0.01cm} \textcolor{green}{\faCheckCircle}  \vspace{0.1cm}& 
         \vspace{0.01cm} \textcolor{green}{\faCheckCircle} \vspace{0.1cm}& 
         \vspace{0.01cm} \textcolor{green}{\faCheckCircle} \vspace{0.1cm}& 
         \vspace{0.01cm} \textcolor{red}{\faTimesCircle} \vspace{0.1cm}& 
         \vspace{0.01cm} \textcolor{red}{\faTimesCircle} \vspace{0.1cm}& 
         \vspace{0.01cm} \textcolor{red}{\faTimesCircle} \vspace{0.1cm}&
         \vspace{0.01cm} \textcolor{red}{\faTimesCircle} \vspace{0.1cm}&
         \vspace{0.01cm} Survey of different architectures of transformers for image classification, object detection and image segmentation\vspace{0.1cm}\\
         \hline
         \vspace{0.01cm} \cite{Xu2022} \vspace{0.1cm}& 
         \vspace{0.01cm} 2022 \vspace{0.1cm}& 
         \vspace{0.01cm} \textcolor{red}{\faTimesCircle} \vspace{0.1cm}& 
         \vspace{0.01cm} \textcolor{green}{\faCheckCircle} \vspace{0.1cm}& 
         \vspace{0.01cm} \textcolor{green}{\faCheckCircle} \vspace{0.1cm}& 
         \vspace{0.01cm} \textcolor{red}{\faTimesCircle} \vspace{0.1cm}& 
         \vspace{0.01cm} \textcolor{green}{\faCheckCircle} \vspace{0.1cm}& 
         \vspace{0.01cm} \textcolor{red}{\faTimesCircle} \vspace{0.1cm}&
         \vspace{0.01cm} \textcolor{red}{\faTimesCircle} \vspace{0.1cm}&
         \vspace{0.01cm} Survey of transformers in computational visual media \vspace{0.1cm}\\
         
         \hline
         \vspace{0.01cm} Our survey \vspace{0.1cm}& 
         \vspace{0.01cm} 2022 \vspace{0.1cm}& 
         \vspace{0.01cm} \textcolor{green}{\faCheckCircle} \vspace{0.1cm}& 
         \vspace{0.01cm} \textcolor{green}{\faCheckCircle} \vspace{0.1cm}& 
         \vspace{0.01cm} \textcolor{green}{\faCheckCircle} \vspace{0.1cm}& 
         \vspace{0.01cm} \textcolor{green}{\faCheckCircle} \vspace{0.1cm}& 
         \vspace{0.01cm} \textcolor{green}{\faCheckCircle} \vspace{0.1cm}&
         \vspace{0.01cm} \textcolor{green}{\faCheckCircle} \vspace{0.1cm}&
         \vspace{0.01cm} \textcolor{green}{\faCheckCircle} \vspace{0.1cm}&
         \vspace{0.01cm} Survey of applications of transformers in CV, New outlook to the open research gaps \vspace{0.1cm}\\
         \hline
         \hline
     \end{tabular}
     \label{TableSummary}
\end{table*}
\begin{figure}[t]
	\includegraphics[width=0.45\textwidth]{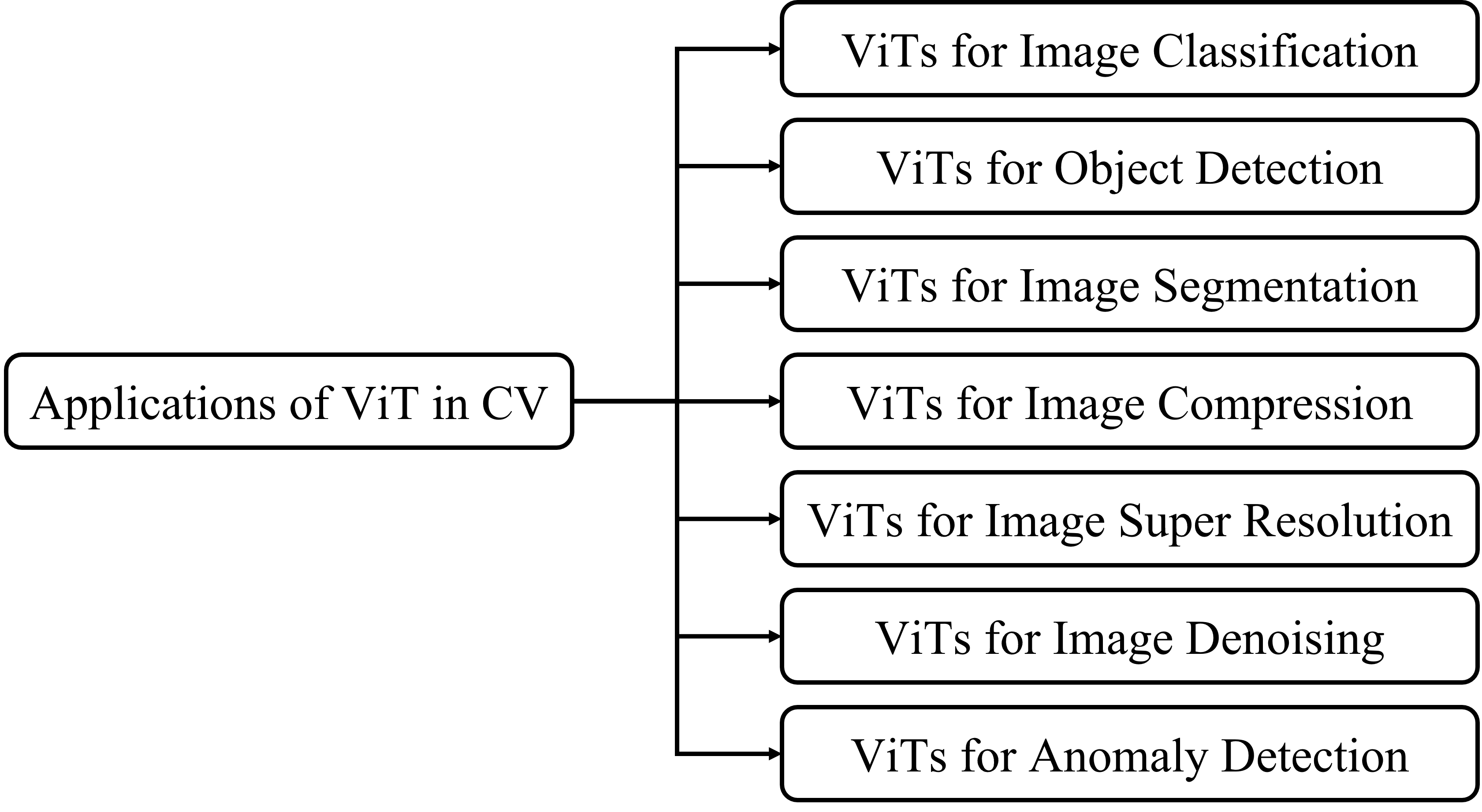}
	\caption{Organization of the section III.}
	\label{Section2}
\end{figure}
\section{Applications of ViTs in CV}
In addition to classical ViTs, modified versions of classical ViTs are used for object detection, image segmentation, compression, super-resolution, denoising, and anomaly detection. Fig. \ref{Section2} shows the organization of section III.

\subsection{ViTs for Image Classification}
In image classification, the image is initially divided into patches; these patches are fed linearly to the transformer encoder, where MLP, normalization, and multi-head attention are applied to create embedded patches. Embedded patches are fed to the MLP head, which predicts the output class. These classical ViTs have been used by many researchers to classify visual objects. 

In \cite{CChen2021}, the authors proposed CrossViT-15, CrossViT-18, CrossViT-9$\dagger$, CrossViT-15$\dagger$, CrossViT-18$\dagger$ for the image classification. They used ImageNet1K, CIFAR10, CIFAR100, pet, crop disease, and ChestXRay8 datasets to evaluate the different variants of CrossViT. They achieved 77.1\% accuracy on the ImageNet1K dataset by using CrossViT-9$\dagger$. Similarly, they attained 82.3\% and 82.8\% accuracy on ImageNet1K dataset using CrossViT-15$\dagger$ and CrossViT-18$\dagger$ respectively. Similarly, the authors got 99.0\% and 99.11\% accuracy with CrossViT-15 and CrossViT-18, respectively, on the CIFAR10 dataset. However, they obtained 90.77\% and 91.36\% accuracy on the CIFAR100 dataset using CrossViT-15 and CrossViT-18, respectively. The authors also used CrossViT for pet classification, crop disease classification, and chest X-ray classification. They observed the highest accuracy of 95.07\% with CrossViT-18 for the pet classification. Similarly, they achieved the highest accuracy of 99.97\% with CrossViT-15 and CrossViT-18 for the crop diseases classification. Moreover, they achieved the highest accuracy of 55.94\% using CrossViT-18 for the chest x-rays classification.

Deng {\em et al.} in \cite{PDeng2022}, proposed a combined CNN and ViT model named CTNet for the classification of high-resolution remote sensing images. To evaluate the model, they used the aerial image dataset (AID) and Northwestern Polytechnical University (NWPU)-RESISC45 dataset. CTNet obtained an accuracy of 97.70\% and 95.49\% using AID and NWPU-RESISC45 datasets, respectively. In \cite{KXu2022}, the authors proposed excellent teacher guiding small networks (ES-GSNet) for the classification of the remote sensing image scenes. They used four datasets: AID, NWPU-RESISC45, UC-Mered Land use dataset (UCM), and OPTIMAL-31. They gained the accuracy of 96.88\%, 94.50\%, 99.29\%, and 96.45\% for AID, NWPU-RESISC45, UCM, and OPTIMAL-31 datasets, respectively. 

Yu {\em et al.} in \cite{SYu2021}, presented multiple instance enhanced ViT (MIL-ViT) for fundus image classification. They used APTOS2019 blindness detection and the 2020 retinal fundus multi-disease image dataset (RFMiD2020). MIL-ViT gave an accuracy of 97.9\% on the APTOS2019 dataset and 95.9\% on the RFMiD2020 dataset.

Xue {\em et al.} in \cite{ZXue2022}, proposed deep hierarchical ViT (DHViT) for the hyperspectral and light detection and ranging (LiDAR) data classification. The authors used Trento, Houston 2013, and Houston 2018 datasets and obtained the accuracy of 99.58\%, 99.55\%, and 96.40\%, respectively.

In \cite{MKaselimi2022} elaborated on the use ViT for the satellite imagery multilabel classification and proposed ForestViT. ForestViT demonstrated the accuracy of 94.28\% on planet understanding of the amazon from space (PUAS) dataset.

In \cite{YChen2022}, the researchers put forward the concept of LeViT for pavement image classification. They used Chinese asphalt pavement and German asphalt pavement to evaluate the model's performance. They obtained an accuracy of 91.56\% using the Chinese asphalt pavement dataset and 99.17\% using the German asphalt pavement dataset.

In \cite{Jamil2022}, the authors used ViT to distinguish malicious drones from the aeroplanes, birds, drones, and helicopters. They demonstrated the efficiency of ViT for the classification over several CNNs such as AlexNet \cite{AlexNet}, ResNet-50 \cite{ResNet2016}, MobileNet-V2 \cite{Mobilenet2018}, ShuffleNet \cite{Shufflenet}, SqueezeNet \cite{Squeezenet}, and EfficicentNetb0 \cite{Efficientnet}. ViT model achieved 98.3\% accuracy on malicious drones dataset.

Tanzi {\em et al.} in \cite{Tanzi2022} applied ViT for the classification of the femur fracture. They used a dataset of real x-rays. The model achieved the accuracy of 83\% with 77\% precision, 76\% recall, and 77\% F1-score.

In \cite{JChen2022}, the authors modified classical ViT and proposed SeedViT for the classification of the maize seeds' quality. They used a custom dataset. The model outperformed CNNs and achieved 96.70\% accuracy. 

Similarly, in \cite{ZWang2022}, the researchers put forward double output ViT (DOViT) for the classification of air quality and its measurement. They used two datasets named get AQI in one shot-1 (GAOs-1) and get AQI in one shot-2 (GAOs-2). The model achieved 90.32\% accuracy for GAOs-1 dataset and 92.78\% accuracy for GAOs-2 dataset.

In \cite{ZSha2022}, the authors developed a novel multi-instance ViT called MITformer for the remote sensing scene classification. They evaluated their model on three different datasets. The model achieved 99.83\% accuracy for the UCM dataset, 97.96\% accuracy for the AID dataset, and 95.93\% accuracy for the NWPU dataset.

Table \ref{Table1} shows the summary of the application of ViT for image classification.
\begin{table*}
    \centering
     \caption{ViT for Image Classification.}
    \begin{tabular}{p{40pt} p{80pt} p{100pt} p{150pt} p{70pt}}
    \hline
    \hline
    \vspace{0.01cm} \textbf{Research} \vspace{0.1cm}& 
    \vspace{0.01cm} \textbf{Model} \vspace{0.1cm}& 
    \vspace{0.01cm} \textbf{Dataset} \vspace{0.1cm}& 
    \vspace{0.01cm} \textbf{Objective} \vspace{0.1cm}&
    \vspace{0.01cm} \textbf{Accuracy} \vspace{0.1cm}\\
    \hline
    \hline
    \multirow{13}{*}{\cite{CChen2021}}& 
    CrossViT-9$\dagger$& 
    \multirow{3}{*}{ImageNet1K}& 
    \multirow{3}{*}{Image classification }&
    77.1\%\\
    & 
    CrossViT-15$\dagger$&
    & 
    & 
    82.3\%\\
    & 
    CrossViT-18$\dagger$& 
    & 
    & 
    82.8\%\\ \cline{2-5}
    & 
      \multirow{5}{*}{CrossViT-15}& 
      CIFAR10& 
      \multirow{2}{*}{Image classification}
    & 
    99.0\%\\
    & 
    & 
    CIFAR100& 
    
    & 
    90.77\%\\ \cline{3-5}
    & 
    & 
    Pet&
    Pet classification& 
    94.55\%\\
    & 
    & 
    Crop Diseases& 
    Crop diseases classification& 
    99.97\% \\
    & 
    & 
    ChestXRay8& 
    Chest X rays classification& 
    55.89\%\\
    \cline{2-5}
    & 
      \multirow{5}{*}{CrossViT-18}& 
      CIFAR10
    & 
      \multirow{2}{*}{Image classification}
    & 
    99.11\%\\
    & 
    & 
    CIFAR100
    & 
    
    & 
    91.36\%\\ \cline{3-5}
    & 
    & 
    Pet
    & 
    Pet classification
    & 
    95.07\%\\
    & 
    & 
    Crop Diseases
    & 
    Crop diseases classification 
    & 
    99.97\%\\
    & 
    & 
    ChestXRay8
    & 
    Chest X rays classification
    & 
    55.94\%\\
    \hline
    
    \multirow{2}{*}{\cite{PDeng2022}}& 
    \multirow{2}{*}{CTNet}& 
    AID& 
    \multirow{2}{*}{Remote sensing scene classification}& 
    97.70\%\\
    & 
    & 
    NWPU-RESISC45& 
    & 
    95.49\%\\
    \hline
     \multirow{4}{*}{\cite{KXu2022}}& 
    \multirow{4}{*}{ET-GSNet}& 
    AID& 
    \multirow{4}{*}{Remote sensing image scene classification}& 
    96.88\%\\
    & 
    & 
    NWPU-RESISC45& 
    & 
    94.50\%\\
    & 
    & 
    UCM& 
    & 
    99.29\%\\
    & 
    & 
    OPTIMAL-31& 
    & 
    96.45\%\\
    \hline
    \multirow{2}{*}{
    \cite{SYu2021}}
    & 
    \multirow{2}{*}{MIL-ViT}
    &
    APTOS2019& 
    \multirow{2}{*}{
    Fundus image classification}& 
    97.9\%\\
    & 
    &
    RFMiD2020& 
    &
    95.9\% \\
    \hline
    \multirow{3}{*}{\cite{ZXue2022}}& 
    \multirow{3}{*}{DHViT}& 
    Trento& 
    \multirow{3}{*}{Hyperspectral and LiDAR data classification}& 
    99.58\%\\
    & 
    & 
    Houston 2013& 
    & 
    99.55\%\\
    & 
    & 
    Houston 2018& 
    & 
    96.40\%\\
    \hline
    \cite{MKaselimi2022}& 
    ForestViT& 
    PUAS& 
    Satellite imagery multilabel classification & 
    94.28\%\\
    \hline
    \multirow{2}{*}{\cite{YChen2022}}& 
    \multirow{2}{*}{LeViT}& 
    Chinese asphalt pavement& 
    \multirow{2}{*}{Pavement image classification}& 
    91.56\%\\
    & 
    & 
    German asphalt pavement& 
    & 
    99.17\%\\
    \hline
    \cite{Jamil2022}& 
    ViT& 
    Malicious drone& 
    Malicious drones classification& 
    98.3\%\\
     \hline
    \cite{Tanzi2022}& 
    ViT& 
    Real X Rays& 
    Femur fracture classification& 
    83.00\%\\
    \hline
   \cite{JChen2022}& 
    SeedViT& 
    Maize seeds& 
    Maize seeds quality classification& 
    96.70\%\\
    \hline
    \multirow{2}{*}{\cite{ZWang2022}}& 
    \multirow{2}{*}{DOViT}& 
    GAOs-1& 
    \multirow{2}{*}{Air quality classification}& 
    90.32\%\\
    & 
    & 
    GAOs-2& 
    & 
    92.78\%\\
    \hline
    \multirow{3}{*}{\cite{ZSha2022}}& 
    \multirow{3}{*}{MITformer}& 
    UCM& 
    \multirow{3}{*}{Remote sensing scene classification}& 
    99.83\%\\
    & 
    & 
    AID& 
    & 
    97.96\%\\
    & 
    & 
    NWPU& 
    & 
    95.93\%\\
    \hline
    \hline
    \end{tabular}
    \label{Table1}
\end{table*}

\subsection{ViTs for Object Detection}
The effort to tame pre-trained vanilla ViT for object detection has never stopped since the evolution of transformer \cite{Vaswani2017} to CV \cite{Dosovitskiy2020}. Beal {\em et al.} \cite{Beal2020} is the first to use a faster region-based convolutional neural network (R-CNN) detector with a supervised pre-trained ViT for object detection. You only look at one sequence (YOLOS) \cite{Fang2021} suggests using simply a pre-trained ViT encoder to conduct object detection in a pure sequence-to-sequence way. Li {\em et al.} \cite{Li2021} are the first to do a large-scale study of vanilla ViT on object detection using sophisticated masked image modeling (MIM) pre-trained representations \cite{Bao2022}-\cite{He2022}, confirming vanilla ViT's promising potential and capacity in object-level recognition.

In \cite{Horvath2021}, the authors proposed the unsupervised learning-based technique using ViT for the detection of the manipulation in the satellite images. They used two different datasets for the evaluation of the framework. The ViT model with post-processing (ViT-PP) achieves an F1-score of 0.354 and a Jaccord index (JI) of 0.275 for dataset 2. F1-score and JI can be calculated by \ref{e1} and \ref{e2} respectively.
\begin{equation}
    F_1=\frac{2*T_P}{2*T_P+F_P+F_N},
    \label{e1}
\end{equation}
\begin{equation}
    JI=\frac{T_P}{T_P+F_P+F_N},
    \label{e2}
\end{equation}
Here $T_P$, $F_P$ and $F_N$ denote true positive, false positive and false negative respectively. 

In \cite{YWang2022}, the authors proposed bridged transformer (BrT) for the 3D object detection. The model was applied for the vision and point cloud 3D object detection. They used ScanNet-V2 \cite{scannet} and SUN RGB-D \cite{sunrgbd} datasets to validate their model. The model demonstrated the mean average precision (mAP)@0.5 of 52.8 for the ScanNet-V2 dataset and 55.2 for the SUN RGB-D dataset. 

Similarly, in \cite{ZLiu2021}, the authors proposed a transformers-based framework for the detection of the 3D objects using point cloud data. They used ScanNet-V2 \cite{scannet} and SUN RGB-D \cite{sunrgbd} datasets to validate their model. The model demonstrated the mean average precision (mAP)@0.5 of 52.8 for the ScanNet-V2 dataset and 45.2 for the SUN RGB-D dataset.  

Table \ref{Table2} shows the application of ViT for the object detection.
\begin{table*}
    \centering
     \caption{ViT for Object Detection.}
    \begin{tabular}{p{30pt} p{45pt} p{100pt} p{165pt} p{70pt} p{30pt}}
    \hline
    \hline
    \vspace{0.01cm} \textbf{Research} \vspace{0.1cm}& 
    \vspace{0.01cm} \textbf{Model} \vspace{0.1cm}& 
    \vspace{0.01cm} \textbf{Dataset} \vspace{0.1cm}& 
    \vspace{0.01cm} \textbf{Objective} \vspace{0.1cm}&
    \vspace{0.01cm} \textbf{Performance metric} \vspace{0.1cm}&
    \vspace{0.01cm} \textbf{Value} \vspace{0.1cm}\\
    \hline
    \hline
    \vspace{0.01cm} \cite{Fang2021} \vspace{0.1cm}& 
    \vspace{0.01cm} YOLOS \vspace{0.1cm}& 
    \vspace{0.01cm} COCO \vspace{0.1cm}& 
    \vspace{0.01cm} Object detection \vspace{0.1cm}& 
    \vspace{0.01cm} $AP^{box}$ \vspace{0.1cm}& 
    \vspace{0.01cm} 42.0 \vspace{0.1cm}\\
    \hline 
    \multirow{2}{*}{\cite{Horvath2021}}& 
    \multirow{2}{*}{ViT}&
    \multirow{2}{*}{Satellite images (dataset 2)}& 
    \multirow{2}{*}{Manipulation detection}& 
    F1-score& 
    0.354\\
    & 
    & 
    & 
    & 
    JI& 
    0.275\\
    \hline 
    \multirow{2}{*}{\cite{YWang2022} }& 
    \multirow{2}{*}{BrT}&
    ScanNet-V2& 
    \multirow{2}{*}{3D object detection}& 
    \multirow{2}{*}{mAP@0.5}& 
    55.2 \\
    & 
    & 
    SUN RGB-D& 
    & 
    & 
    48.1\\
    \hline 
    \multirow{2}{*}{\cite{ZLiu2021}}& 
    \multirow{2}{*}{ViT based}& 
    ScanNet-V2& 
    \multirow{2}{*}{3D object detection using point cloud data}& 
    \multirow{2}{*}{mAP@0.5}& 
    52.8\\
    & 
    & 
    SUN RGB-D& 
    & 
    & 
    45.2\\
    \hline 
    \hline
    \end{tabular}
    \label{Table2}
\end{table*}

\subsection{ViTs for Image Segmentation}
Image segmentation can also be done using transformers. A combination of ViT and U-Net was used in ~\cite{JChen2021} to segment medical images. The authors replaced the encoder part of the classical U-Net with a transformer. A multi-atlas abdomen labeling challenge dataset from MICCAI 2015 was used. By using images of resolution 224, the TransUNet achieved an average dice score of 77.48\%, while using images of resolution 512, it achieved an average dice score of 84.36\%.

In~\cite{ASagar2022} the authors proposed a "ViT for biomedical image segmentation (ViTBIS)" for medical image segmentation. Transformers were used for both encoders and decoders in their transformer-based model. In addition, the MICCAI 2015 multi-atlas abdomen labeling challenge dataset and the Brain Tumor Segmentation (BraTS 2019) challenge dataset were used. The evaluation metric used was dice score and Hausdorff distance (HD)~\cite{HD1993}. According to the MICCAI 2015 dataset, average dice scores were 80.45\%, and average HDs were 21.24\%.

Hatamizadeh {\em et al.} in \cite{Hatamizadeh2022}, proposed UNetFormer for the medical image segmentation. The model contained a transformer-based encoder, decoder, and bottleneck part. They used  medical segmentation decathlon (MSD) \cite{MSD} and BraTS 2021 \cite{BraTS2021} dataset to test UNetFormer. They evaluated dice scores and HD. The dice score using the MSD dataset was 96.03\% for liver and 59.16\% for tumor, whereas the value of HD was 7.21\% for liver and 8.49\% for tumor. Moreover, the average dice score on the BraTS 2021 dataset is 91.54\%.  

In \cite{ZYang2021}, the authors proposed a novel ``language-aware ViT (LAVT)'' for the image segmentation. They used four different datasets for the evaluation of the model. The datasets were RefCOCO \cite{RefCOCO}, RefCOCO+ \cite{RefCOCO}, G-Ref (UMD partition) \cite{GRefU} and G-Ref (Google partition) \cite{GRefG}. They used intersection over union (IoU) as the performance metric. The value of IoU for the RefCOCO dataset was 72.73\%, and for RefCOCO+, the IoU was 62.14\%. Similarly, for G-Ref (UMD partition), IoU was 61.24\%, and for G-Ref (Google partition), IoU was 60.50\%.

Table \ref{Table3} shows the application of ViT for the image segmentation.
\begin{table*}
    \centering
     \caption{ViT for Image Segmentation.}
    \begin{tabular}{p{35pt} p{75pt} p{100pt} p{120pt} p{70pt} p{30pt}}
    \hline
    \hline
    \vspace{0.01cm} \textbf{Research} \vspace{0.1cm}& 
    \vspace{0.01cm} \textbf{Model} \vspace{0.1cm}& 
    \vspace{0.01cm} \textbf{Dataset} \vspace{0.1cm}& 
    \vspace{0.01cm} \textbf{Objective} \vspace{0.1cm}&
    \vspace{0.01cm} \textbf{Performance metric} \vspace{0.1cm}&
    \vspace{0.01cm} \textbf{Value} \vspace{0.1cm}\\
    \hline
    \hline
    \cite{JChen2021} & 
    TransUNet & 
    MICCAI 2015 & 
    Medical image segmentation & 
    Dice score & 
    77.48\%\\
    \hline 
    \multirow{2}{*}{\cite{ASagar2022} }& 
    \multirow{2}{*}{ViTBIS}& 
    MICCAI 2015& 
    \multirow{2}{*}{Medical image segmentation}& 
    Dice score& 
    80.45\%\\
    & 
    & 
    & 
    & 
    HD& 
    21.24\%\\
    \hline
    \multirow{5}{*}{ \cite{Hatamizadeh2022}}& 
    \multirow{5}{*}{UNetFormer}& 
    \multirow{4}{*}{MSD}&
    \multirow{2}{*}{Liver segmentation}& 
    Dice score& 
    96.03\%\\
    & 
    & 
    & 
    & 
    HD& 
    7.21\%\\ \cline{4-6}
    & 
    & 
    & 
    \multirow{2}{*}{Tumor segmentation}& 
    Dice score& 
    59.16\%\\
    & 
    & 
    & 
    & 
    HD & 
    8.49\% \\ \cline{3-6}
    & 
    & 
    BraTS 2021 & 
    Brain tumor segmentation & 
    Dice score & 
    91.54\% \\
    \hline
    \multirow{4}{*}{\cite{ZYang2021}}& 
    \multirow{4}{*}{LAVT}&
    RefCOCO& 
    \multirow{4}{*}{Image segmentation}& 
    \multirow{4}{*}{IoU}&
    72.73\%\\
    & 
    & 
    RefCOCO+ & 
    & 
    & 
    62,14\%\\
    & 
    & 
     G-Ref (UMD partition) & 
    & 
    & 
    61.24\% \\
    & 
    & 
    G-Ref (Google partition) & 
    & 
    & 
    60.50\%\\
    \hline
    \hline
    \end{tabular}
    \label{Table3}
\end{table*}

\subsection{ViTs for Image Compression}
In recent years, learning-based image compression has been the focus of research. For lossy image compression based on learning, different CNN-based architectures proved effective. As ViTs evolved, learning-based image compression was also done by transformer-based models. In \cite{YQian2022}, the authors modified the entropy module of the Ball{\' e} 2018 mode \cite{Balle2018} with the ViT. Due to the fact that the entropy module used a transformer, this model was called Entroformer. Entroformer effectively captured long-range dependencies in probability distribution estimation. On the Kodak dataset, they demonstrated the performance of the Entroformer. When the model was optimized for the mean squared error (MSE) loss function, the average peak signal-to-noise ratio (PSNR) and multi-scale structural similarity (MS-SSIM) were 27.63 dB and 0.90132, respectively.

\subsection{ViTs for Image Super Resolution}
CNN has been used to perform image super-resolution. With ViT's superiority over CNN, image super-resolution can also be achieved by transformers. Spatio-temporal ViT, a transformer-based model for super-resolution of microscopic images, is developed in~\cite{Christensen2022}. Additionally, the model addressed the problem of video super-resolution. To test the model's performance, the authors used a video dataset. PSNR was calculated for static and dynamic videos. Static, medium, fast, and extreme motions were considered. The PSNR for static was 34.74 dB, whereas the PSNR for medium, fast and extreme was 30.15 dB, 26.04 dB, and 22.95 dB, respectively. 

\subsection{ViTs for Image Denoising}
Denoising images has also been a challenging problem for researchers. In spite of this, ViT has found a solution. A transformer was used to denoise CT images in~\cite{TEDNet}. They proposed a model called TED-Net for low-dose CT denoising. The authors used a transformer for both the encoder and decoder parts. Using the AAPM-Mayo clinic LDCT Grand Challenge dataset, they obtained structural similarity (SSIM) of 0.9144 and root mean square error (RMSE) of 8.7681.

\begin{table*}
    \centering
     \caption{ViT for Anomaly Detection.}
    \begin{tabular}{p{35pt} p{110pt} p{50pt} p{130pt} p{70pt} p{30pt}}
    \hline
    \hline
    \vspace{0.01cm} \textbf{Research} \vspace{0.1cm}& 
    \vspace{0.01cm} \textbf{Model} \vspace{0.1cm}& 
    \vspace{0.01cm} \textbf{Dataset} \vspace{0.1cm}& 
    \vspace{0.01cm} \textbf{Objective} \vspace{0.1cm}&
    \vspace{0.01cm} \textbf{Performance metric} \vspace{0.1cm}&
    \vspace{0.01cm} \textbf{Value} \vspace{0.1cm}\\
    \hline
    \hline
    \multirow{3}{*}{\cite{PMishra2021}}& 
    \multirow{3}{*}{VT-ADL}& 
    MNIST & 
    \multirow{3}{*}{Anomaly detection and localization}&
    \multirow{3}{*}{PRO}&
    0.984\\
    & 
    & 
    MVTec& 
    & 
    & 
    0.807\\
    & 
    & 
    BTAD& 
    & 
    & 
    0.89\\
    \hline 
    \multirow{3}{*}{\cite{YLee2022}}& 
    \multirow{3}{*}{AnoViT}& 
    MNIST& 
    \multirow{3}{*}{Anomaly detection and localization}&
    \multirow{3}{*}{AUROC}& 92.4\%\\
    & 
    & 
    CIFAR& 
    & 
    & 
    60.1\%\\
    
    & 
    & 
    MVTec & 
    & 
    & 
    78.0\% \\
    \hline 
    \multirow{6}{*}{\cite{HYuan2021}}& 
    \multirow{3}{*}{TransAnomaly (without swm)}& 
    Pred1& 
    \multirow{6}{*}{Anomaly detection in videos}& 
    \multirow{6}{*}{AUC}&
    84.00\%\\
    
    & 
    & 
    Pred2& 
    & 
    & 
    96.10\%\\
    
    & 
    & 
    Avenue& 
    & 
    & 
    85.80\%\\ \cline{2-3} \cline{6-6}
    & 
    \multirow{3}{*}{TransAnomaly (with swm)}& 
    Pred1& 
    & 
    & 
    86.70\%\\
    & 
    & 
    Pred2& 
    & 
    & 
    96.40\%\\
    & 
    & 
    Avenue& 
    & 
    & 
    87.00\%\\
    \hline 
    \hline
    \end{tabular}
    \label{Table7}
\end{table*}
\subsection{ViTs for Anamoly Detection}
Additionally, ViT is used for anomaly detection. A novel ViT network for image anomaly detection and localization (VT-ADL) was developed in~\cite{PMishra2021}. In their study, the authors used a real-world dataset called BTAD. The model was also tested on two publicly available datasets, MNIST and MVTec \cite{mvtecdataset}. For all three datasets, they calculated the model's per region overlap (PRO) score. A mean PRO score of 0.984 was obtained for the MNIST dataset, 0.807 for the MVTec dataset, and 0.89 for the BTAD dataset. 

Similarly, in \cite{YLee2022}, the authors proposed AnoViT for the detection and localization of anomalies. MNIST, CIFAR, and MVTecAD datasets were used by the authors. Based on the MINST, CIFAR, and MVTecAD datasets, the mean area under region operating characteristics (AUROC) curve was 92.4, 60.1, and 78, respectively. 

Yuan {\em et al.} in \cite{HYuan2021} proposed TransAnomaly, a video ViT and U-Net-based framework for the detection of the anomalies in the videos. They used three datasets, Pred1, Pred2, and Avenue. The calculated area under the curve (AUC) for three datasets achieved 84.0\%, 96.10\%, and 85.80\%, respectively, without using the sliding windows method (swm). With the swm, the model gave AUC of 86.70\%, 96.40\%, and 87.00\% for the Pred1, Pred2, and Avenue datasets, respectively.

Table \ref{Table7} shows the summary of the ViT for anomaly detection.

\section{Open Research Challenges and Future Directions}
Despite showing promising results for different image coding and CV tasks. In addition to high computational costs, large training datasets, neural architecture search, interpretability of transformers, and efficient hardware designs, ViTs implementation still faces challenges. The purpose of this section is to explain the challenges and future directions of ViTs.
\subsection{High computational cost}
There are millions of parameters in ViT-based models. Computers with high computational power are needed to train these models. Due to their high cost, these high-performance computers increase the computational cost of ViTs. In comparison to CNN, ViT performs better; however, its computational cost is much higher. One of the biggest challenges researchers face is reducing the computational cost of ViTs.

\subsection{Large training dataset}
ViTs' training requires a large amount of data. With a small training dataset, ViTs perform poorly. ViT trained with the ImageNet1K dataset performs worse than ResNet, but ViT trained with ImageNet21K performs better than ResNet. 

\subsection{Neural architecture search (NAS)}
There has been a great deal of exploration of NAS for CNNs. In contrast, NAS has not yet been explored for ViTs. The NAS exploration for ViTs gives young investigators a new direction for the future.

\subsection{Interpret-ability of the transformers}
It is difficult to visualize the relative contribution of input tokens to the final predictions with ViTs since the attention that originates in each layer is intermixed in succeeding layers. The problem remains unresolved.

\subsection{Hardware efficient desings}
Power and processing requirements can make large-scale ViTs networks unsuitable for edge devices and resource-constrained contexts such as the internet of things (IoT).

\section{Conclusion}
It is becoming more common to use ViTs for image coding and CV instead of CNNs. The use of ViTs for classification, detection, segmentation, compression, and image super-resolution has risen dramatically since the introduction of the classical ViT for image classification. This survey presented the existing surveys on ViTs in the literature. This survey highlighted the applications of different variants of ViTs in CV. This survey examined the use of ViTs for image classification, object detection, image segmentation, image compression, image super-resolution, image denoising, and anomaly detection. We also presented the lessons learned in each category. Additionally, we discussed the open research challenges faced by the researchers during the implementation of ViTs, such as high computational costs, large training datasets, interpretability of transformers, and hardware efficiency. By providing future directions, we gave the young researchers a new perspective.

\ifCLASSOPTIONcaptionsoff
  \newpage
\fi

%








\end{document}